# Improving Data Quality Through Deep Learning and Statistical Models

Wei Dai[*1], Kenji Yoshigoe[#2], William Parsley[*3],
{Information Science[*], Computer Science[#]}, University of Arkansas at Little Rock,
Little Rock, AR, USA
{wxdai[1], kxyoshigoe[2], wmparsley[3]}@ualr.edu

## Abstract

*Outlier detection is an essential aspect of data quality control as it allows analysts and engineers the ability to identify data quality problems through the use of their own data as a tool. However, traditional data quality control methods are based on users' experience or previously established business rules, and this limits performance in addition to being a very time consuming process and low accuracy. Utilizing big data, we can leverage computing resources and advanced techniques to overcome these challenges and provide greater value to the business.*

*In this paper, we first review relevant works and discuss machine learning techniques, tools, and statistical quality models. Second, we offer a creative data profiling framework based on deep learning and statistical model algorithms for improving data quality. Third, authors use public Arkansas officials' salaries, one of the open datasets available from the state of Arkansas' official website, to demonstrate how to identify outlier data for improving data quality via machine learning. Finally, we discuss future works.*

## 1. Introduction

In the big data era, engineers are challenged to innovate software platforms to store, analyze, and manage vast amounts of various datasets. Without improved data quality at data warehouses or data centers, however, we are not able to properly analyze this data due to garbage in, garbage out (GIGO) computing behavior. Hence, improving and controlling data quality has become more and more key business strategies for many companies, organizations, governments, and financial services providers. One of the important tasks for data quality is to detect data quality problems, especially outlier detection.

Dr. Richard Wang, a MIT professor, mentions a theoretical data quality framework. This framework should be representational, intrinsic, contextual, and easily accessible [1]. However, his paper does not go into much detail on how one might recognize, identify, and discover data quality problems.

Popular data profiling tools contain many data profiling modules, but these functions usually are based on simple statistics algorithms (see Fig 1.) [2]. In addition, these tools often do not offer insight into how to improve the data quality problems being identified via machine learning algorithms.

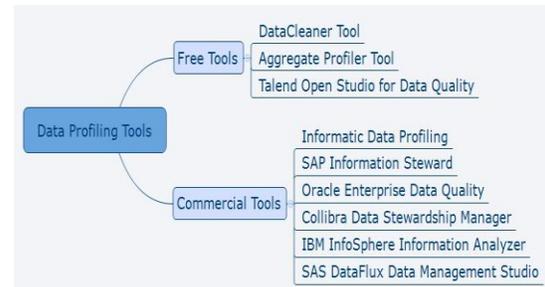

Figure 1. Popular Data Profiling Tools [2]

Machine learning is a promising subfield of computer science to approach human-level artificial intelligence (AI), and is already used to self-driving car, data mining, and natural speech recognition [3-6]. The AI algorithms can either be unsupervised or supervised. Additionally, deep learning is one of the supervised machine learning algorithms [7-9]. Improving data quality is essential to data mining, data analysis, and big data. [16] mentions data quality assessment architecture via AI, but this paper does not discuss how to improve and how to identify data quality with special details based on AI and statistical modes. To the best of our knowledge, no literature improves data quality through both machine learning and statistical quality control models.

The rest of this paper is organized as follows. Section 2 is related work; Section 3 discusses software architecture and data flow; Section 4 describes the data source and data structure used in this work; Section 5 describes data preparation; and Section 6 discusses machine learning modes, including deep learning and statistical quality controls. Section 7 is a summary of our work, and Section 8 is future discussion.

## 2. Related Work
### 2.1 Outlier Detection Technologies

Outlier data means this data is totally different from the others. Hawkins officially defined it as "An outlier is an observation which deviates so much from the other observations as to arouse suspicions that it was generated by a different mechanism"[10]. However, outlier data does not mean error data or problem data, but this data means it has potential error data or risk data. Outlier detections could be used in different industries. For example, [11] mentions that intrusion detection systems, credit card fraud, interesting sensor events, medical diagnosis, law enforcement, and earth science can utilize this technology.

According to [11], there are many outlier detections helping people to identify outlier data, including probabilistic and statistical models, linear correlation analysis, proximity-based detection, and supervised outlier detection.

### 2.2 Statistical Quality Control

Probabilistic and statistical models can be used for outlier detection, including normal distribution and Zipf distribution. Moreover, these models work well in one-dimensional and multidimensional data.

Statistical process control (SPC) [12-14] is a procedure of quality control, and it is based on statistical methods. A normal distribution is a symmetric, continuous, bell-shaped distribution of a variable. Fig. 2 displays a normal probability distribution as a model for a quality characteristics with the specification limits at three or six standard deviations ($\sigma$) on either side of the mean ($\mu$).

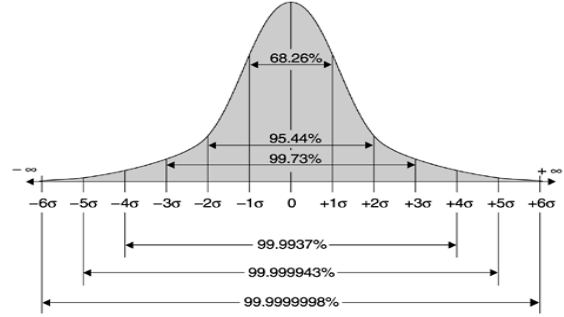

Figure 2. Areas under a Normal Distribution Curve

The base of SPC is the central limit theorem [14]. In addition, a control chart is one of the key techniques of SPC, and this chart contains a center line (CL) and lower and upper control limits (LCL and UCL in formula 1 and 2).

$$UCL = \mu + 3\sigma \quad \text{(Formula 1)}$$
$$LCL = \mu - 3\sigma \quad \text{(Formula 2)}$$

where $\mu$ is population mean, and $\sigma$ is population standard deviation.

A control chart is a graph or chart with limit lines, or control lines. There are basically three kinds of control lines: UCL, CL, and LCL (see Fig. 3).

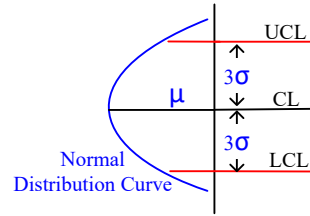

Figure 3. Quality Control Chart of SPC

### 2.3 Deep Learning

Deep learning is a computational model made up of multiple hidden layers of analysis used to understand representations of data so that analysts are able to better understand complex data problems (see Fig. 4). Deep learning network includes Convolutional neural networks and Backpropagation networks.

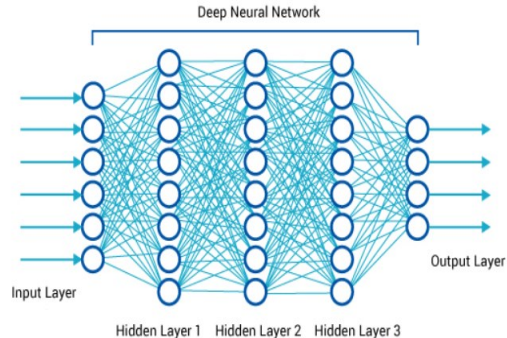

Figure 4. Deep Learning Network

In this paper, we develop deep learning software based on KNIME and WEKA [16-20] because this software are visually interactive tools and the software supports JAVA programs.

## 3. Overview Architecture

In this work, we design a data profiling architecture with three parts: data preparation, machine learning, and data visualization. Data preparation focuses on identifying basic problem and cleaning data. Machine learning contains neutral network and statistical quality control models for discovering complex outlier data (see Fig. 5). Data visitation displays risk data, error data, and correct data.

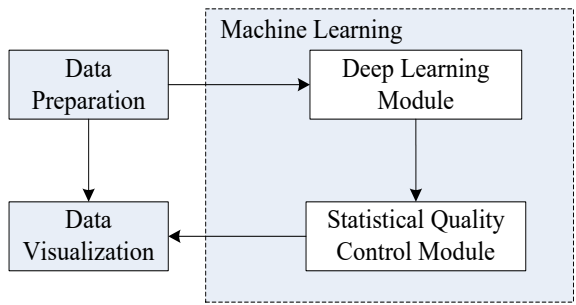

Figure 5. Data Profiling Architecture

Data flow chart (see Fig. 12) shows the procedure of data. At data preparation stage, data profiling program reads data file (CSV file type); string type and date type should be transferred to a numeric value; basically data quality problems should be discovered at this stage, such as incorrect sex code or date. At deep learning network stage, the data is spitted to training data and test data for neural network algorithms. Deep learning network offers predictive salaries. At statistical quality control module stage, the program calculates the errors between predictive salary (P) and real salary (T). If the difference rate is out of UCL or LCL, this data will be marked as outlier data. At data visualization stage, program will produce reports through visualization models, which is not the focus of this work and is omitted from this paper.

## 4. Data Source

For our data source, we access the state of Arkansas' open data library and select their public information sheets containing employee salaries. We download it as a csv file from a website: http://transparency.arkansas.gov. (as shown in Fig. 6).

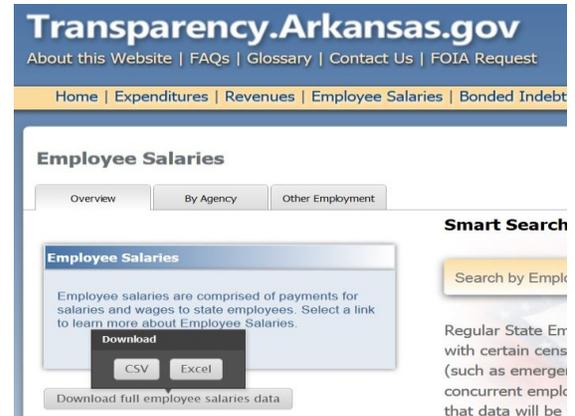

Figure 6 Arkansas State Government Open Datasets

The dataset is open data and contains fiscal year, agency, pay class category, pay scale type, position number, position tile, employee name, percent of time, annual salary, etc.

In this paper, we assume that most of officers' salaries data is accurate and correct because this data comes from Arkansas state human resource department. However, we also believe that this data may contain some errors.

## 5. Data Preparation

Data preparation part contains *Data Transform* and *Data Clean* programs so that basic data quality problems are filtered as shown in Fig. 7. *Data Transform* program is the process of transferring data from String, Boolean, and Character types to numeric types via a hash code function.

*Data Transform* program is written by JAVA program (Fig. 8). There are three major benefits after transferring numeric data.

1) Compress data: String value is too long, but integer value is simple and short. For instance, a hash code of "DEPT OF PARKS AND TOURISM" is 1971843741, resulting in highly efficient compression.
2) Identify problems: a tiny change of string leads to a big different hash code. For example, a hash code of "DEPT OF PARKS AND TOURISM" is 1971843741, but "DEPT OF PARK AND TOURISM" is 1553109818.
3) Machine learning is running well with numeric values.

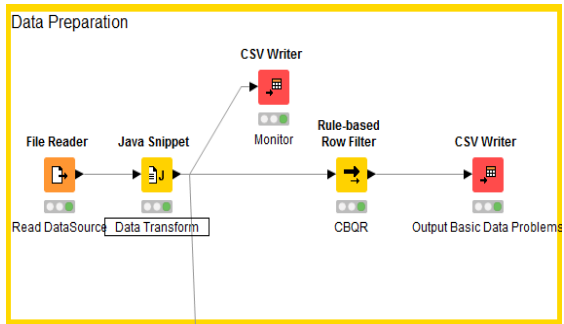

Figure 7 Data Preparation Program

### 5.1 Transferring String Type

If data is null and empty, this data should be marked as a special integrate value for future procedure function. The pseudo-code is here:

```
Read data
IF object is null THEN
    iHashCode = -1;
ELSEIF object is empty THEN
    iHashCode =0;
ELSE
   iHashCode = hashcode(object)
EndIF
```

### 5.2 Transferring Date Type

Another kind of data is date type, such as birthday and career service date. For example, the program transfers birthday date into how many days using the following pseudo-code:

```
Read data
    iFlag = -1
    IF date value is incorrect THEN
        Mark error data, return error
    ELSE
        Calculate days, return iDays
    ENDIF
```

### 5.3 Data Clean

Check Basic Quality Rule (CBQR) means that the program checks the element data quality problems via logical rules. For example, data is empty but it cannot be an empty value. The pseudo-code is here:

```
Read data
    iFlag = -1
    IF  iHashCode == -1 THEN
        IF the data cannot be NULL value THEN
            iFlag =0
        END IF
    ENDIF
    IF iHashCode == 0 THEN
        IF the data cannot be EMPTY value THEN
            iFlag =0
        END IF
    ENDIF

IF iFlag=0 THEN
   Push the data into element quality problem data
EndIF
Close Fetch
```

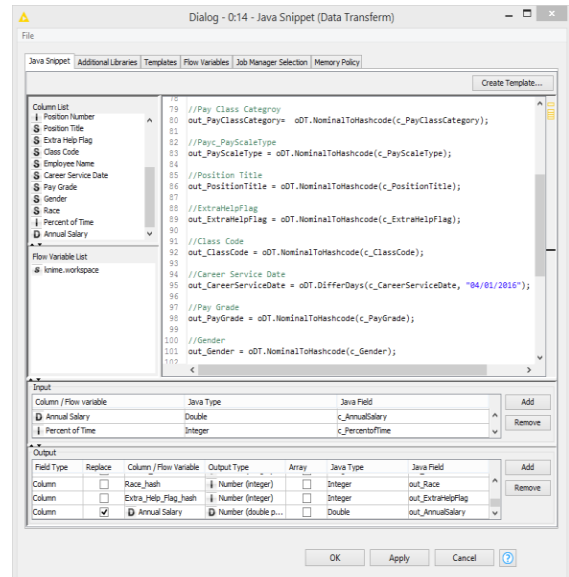

Figure 8 Data Transform Program

## 6.  Deep Learning

*Machine Learning* program is able to learn patterns from data sets. Data should be separated into two parts: training data and testing data. *Machine Learning* program trains itself through training data, and it tests the accuracy of the *Machine Learning* algorithm via testing data. Weka has a test option for choosing different train data sets or test sets. For example, the percent of learning data can be set to 33% as shown in Fig. 9.

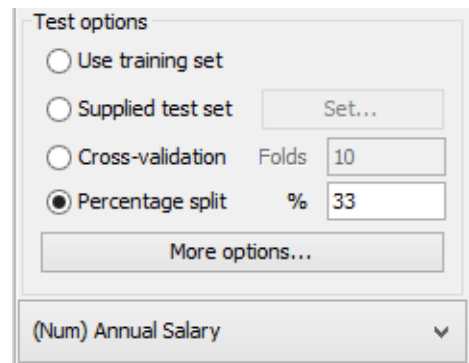

Figure 9 Choosing Data Sets

There are many algorithms for predication. After comparing Backpropagation (BP) network, Convolutional Neural Network (CNN), Support Vector Machine (SVM), and Regression models, testing results show that BP network is the best choice.

As shown in Figure 10, there are six Layers BP network, including 10 input nodes, four hidden layers (each hidden layer has 12, 18, 12, and 10 nodes, respectively), and one output node. Furthermore, Input nodes contain *agency* hashcode, *pay-class-category* hashcode, *pay-scale-type* hashcode, etc. Output node is *annual salary*.

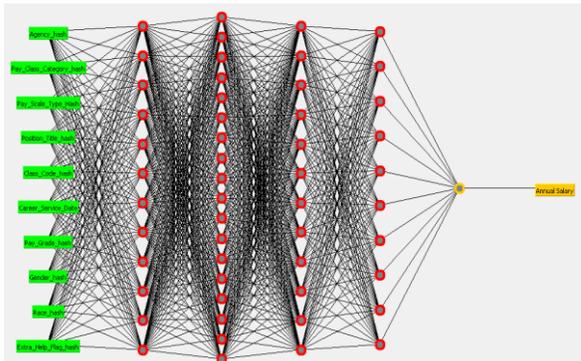

Figure 10. Backpropagation Network

After testing different hidden layers of BP algorithms, we compare test results, including correlation coefficient, mean absolute error, and root mean squared error as shown in Table 1.

According to the table, when hidden layers are 12, 18, 12, and 10, the correlation coefficient has the maximum value. Additionally, mean absolute error and root mean squared error are the minimum value.

TABLE 1. Different hidden layers for BP

| No. | Hidden Layers | correlation coefficient | mean absolute error | root mean squared error |
|---|---|---|---|---|
| 1 | 12,18,12 | 0.7756 | 8381.9728 | 12342.309 |
| 2 | 12,18,12,10 | 0.7846 | 8149.966 | 12133.4712 |
| 3 | 12,18,12,10,10 | 0.0225 | 12602.9016 | 19612.9531 |
| 4 | 12,18,12,10,8 | 0 | 12602.9034 | 19612.9514 |
| 5 | 12,18,24,10 | 0.7768 | 8701.5927 | 12315.6109 |
| 6 | 12,18,24 | 0.7712 | 8445.6757 | 12405.1537 |
| 7 | 12,36,24 | 0.75 | 8605.0589 | 13045.9201 |
| 8 | 12,36,24,10 | 0.7707 | 8424.1713 | 12472.9748 |
| 9 | 24,18,12,10 | 0.7765 | 8362.1652 | 12303.3004 |
| 10 | 12,18,10,10 | 0.7801 | 8300.8978 | 12352.81 |
| 11 | 12,18,16,10 | 0.7774 | 8543.8297 | 12277.3229 |
| 12 | 10,18,12,10 | 0.7787 | 8437.5677 | 12289.3999 |

## 7. Statistical Quality Control Model

Outlier detection is based on statistical quality control mode. Deep learning model outputs a predicated value when it reads input data. However, the predicated value usually is not equal to actual value because of error.

The error for the dataset means the difference between predicated salary and actual salary (Formula 3), and difference ratio is error divides actual salary (Formula 4).

$$Error = Predicated\ Salary - Acutal\ Salary \quad (Formula\ 3)$$

$$Difference\ Ratio = \frac{Error}{Acutal\ Salary} \quad (Formula\ 4)$$

Moreover, quality control chart needs to know mean and standard deviation. However, we can catch the population mean and the standard deviation of the population through the central limit theorem. This theorem has four important properties [15]:

1) The mean of the sample ($\mu_{\bar{x}}$) means will be the same as the population mean ($\mu$).
$$\mu_{\bar{x}} = \mu \quad (Formula\ 5)$$

2) The standard deviation of the sample ($\sigma_{\bar{x}}$) means will be less than the standard deviation of the population ($\sigma$), and it will equal to the population standard deviation divided by the square root of the sample size.
$$\sigma_{\bar{x}} = \frac{\sigma}{\sqrt{n}} \quad (Formula\ 6)$$

3) If the original variable is normally distributed, the distribution of the sample means will be normally distributed, for any sample size n.

4) If the distribution of the original variable might not be normal, a sample size of 30 or more is needed to use a normal distribution to approximate the distribution of the sample means. The larger the sample, the better the approximation will be.

Before calculating quality control chart, we randomly select 100 rows to figure out standard deviation of the sample ($\sigma_{\bar{x}}$) and mean of the sample ($\mu_{\bar{x}}$) for of Difference-Ratio column as a partial sample data are shown in Table 2.

After choosing 100 sample data, we get:
$$\mu_{\bar{x}} = -0.23\ and\ \sigma_{\bar{x}} = 0.22 \quad (Formula\ 7)$$

According to formula 5-7, the population mean ($\mu$) and standard deviation of the population ($\sigma$) are:

$$\mu = \mu_{\bar{x}} = -0.23\ and\ \sigma = \sqrt{n} \times \sigma_{\bar{x}} = \sqrt{100} \times 0.22 = 2.2 \quad (Formula\ 8)$$

Table 2. Part of Sample Data

| ID | Actual AnnualSalary | PredicatedAnnual Salary | ERROR | Difference Ratio |
|---|---|---|---|---|
| 14 | 153588.66 | 89940.368 | -63648.3 | -0.414407496 |
| 49 | 102937.54 | 97946.535 | -4991.01 | -0.048485761 |
| 19 | 95229.36 | 75777.512 | -19451.8 | -0.204263139 |
| 86 | 88607.79 | 87640.989 | -966.801 | -0.010911016 |
| 93 | 69733.87 | 34973.542 | -34760.3 | -0.498471231 |
| 27 | 69733.87 | 32259.939 | -37473.9 | -0.537384932 |
| 63 | 64970.88 | 45202.44 | -19768.4 | -0.304266157 |
| 42 | 61614.18 | 24997.06 | -36617.1 | -0.594296962 |
| 57 | 61566.96 | 32533.174 | -29033.8 | -0.471580634 |
| 75 | 60879.1 | 35386.174 | -25492.9 | -0.418746762 |
| 81 | 57852.08 | 25052.524 | -32799.6 | -0.566955518 |
| 1 | 56264.64 | 50359.792 | -5904.85 | -0.104947761 |
| 66 | 54999.98 | 21391.136 | -33608.8 | -0.611070113 |
| 4 | 50678.99 | 42406.105 | -8272.89 | -0.163240921 |
| 95 | 50607.23 | 19280.438 | -31326.8 | -0.619018113 |
| 13 | 50000.08 | 52144.907 | 2144.827 | 0.042896471 |

According to formula 8, upper and lower control limits (UCL and LCL) are:

$$UCL = \mu + 3\sigma = -0.23 + 3 \times 2.2 = 5.37$$

$$UCL = \mu - 3\sigma = -0.23 - 3 \times 2.2 = -6.83$$

Consequently, the control chart is shown in Figure 11.

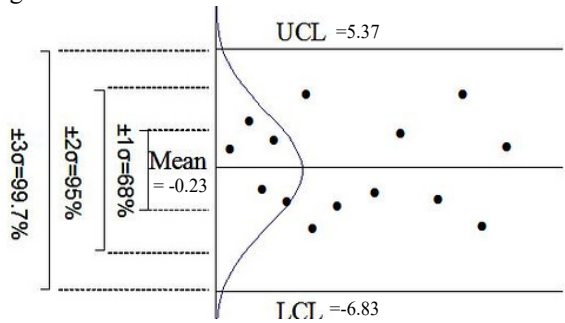

Figure 11 Statistical Quality Control Chart

If difference ratio is out of UCL or LCL, this data is out of three sigma (3σ), and it will be marked as outlier data, or risk data. At Table 3, some data is marked as outlier data because the difference ratio is more than 25. For example, salaries for some full-time employees for Arkansas government were only 21 cents per year.

## 8. Conclusion

Authors have built a unique framework for outlier detection of data profiling, including data preparation, machine learning, and statistical quality control models. After cleaning data at data preparation, machine learning model outputs have predicated salaries. Then, outlier data was detected by statistical quality control model.

Table 3. Outlier Data of Annual Salary

| ID | Actual AnnualSalary | PredicatedAnnual Salary | ERROR | Difference Ratio |
|---|---|---|---|---|
| 14138 | 0.21 | 30607.41 | 30607.20 | 145748.59 |
| 10435 | 0.21 | 26211.88 | 26211.67 | 124817.48 |
| 7238 | 0.21 | 26089.28 | 26089.07 | 124233.67 |
| 1032 | 0.21 | 25926.96 | 25926.75 | 123460.71 |
| 2934 | 0.21 | 25596.23 | 25596.02 | 121885.82 |
| 9102 | 0.21 | 23615.14 | 23614.93 | 112452.06 |
| 2179 | 0.21 | 23167.60 | 23167.39 | 110320.90 |
| 198 | 2040 | 91925.23 | 89885.23 | 44.06 |
| 13131 | 2040 | 91925.23 | 89885.23 | 44.06 |
| 9719 | 2040 | 65939.44 | 63899.44 | 31.32 |
| 11540 | 2040 | 53485.19 | 51445.19 | 25.22 |
| 9998 | 6240 | 22113.57 | 15873.57 | 2.54 |
| 9567 | 5824 | 19029.40 | 13205.40 | 2.27 |
| 17602 | 5824 | 18935.52 | 13111.52 | 2.25 |
| 15575 | 6760 | 21978.68 | 15218.68 | 2.25 |
| 15078 | 5824 | 18474.99 | 12650.99 | 2.17 |
| 2641 | 5824 | 18461.80 | 12637.80 | 2.17 |
| 5988 | 39399.84 | 123755.51 | 84355.67 | 2.14 |
| 2313 | 6240 | 19483.77 | 13243.77 | 2.12 |
| 7521 | 39399.84 | 122850.90 | 83451.06 | 2.12 |
| 15555 | 39399.84 | 122850.90 | 83451.06 | 2.12 |
| 6731 | 7280 | 22524.36 | 15244.36 | 2.09 |

Our work integrates deep learning networks and statistical quality control model for improving data quality. This idea has a great potential to reduce workload and improve performance as system developers and business users do not need to create many data rules to identify data quality problems because deep learning algorithms automatically learn data patterns through training datasets.

## 9. Future Work

Data profiling of information quality is very important for big data. Deep learning is a very promising approach for potentially solving many big data challenges. We will continue to explore other neural network models with various datasets. We will also conduct these experiments on GPU-based server to expedite the deep learning performance, and seek effective data visualization solutions.

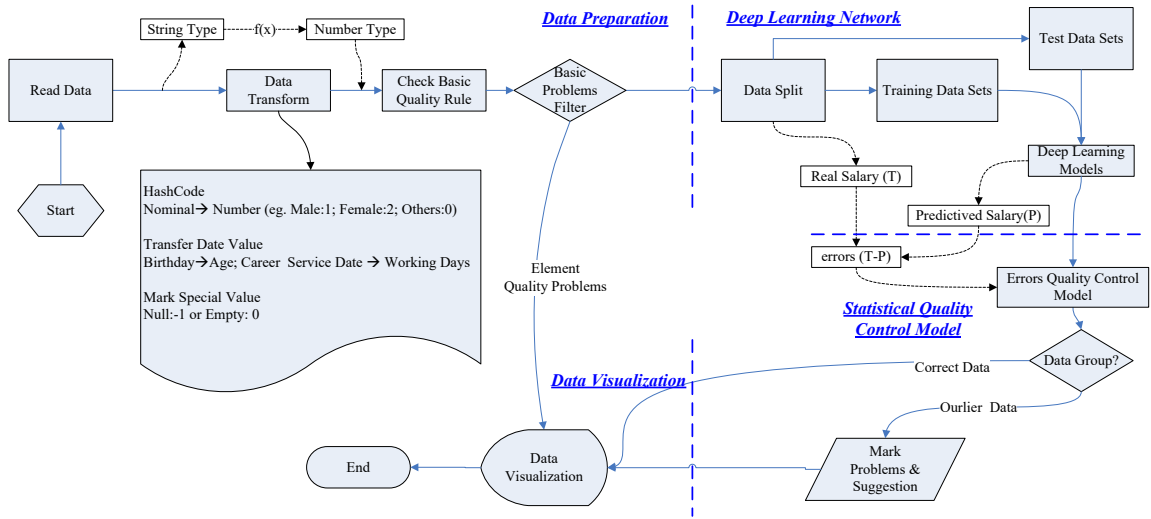

Figure 12 Logical Data Flow Chart